# An Essay Concerning Machine Understanding

Herbert L. Roitblat

ABSTRACT

*Artificial intelligence systems exhibit many useful capabilities, but they appear to lack understanding. This essay describes how we could go about constructing a machine capable of understanding. As John Locke (1689) pointed out words are signs for ideas, which we can paraphrase as thoughts and concepts. To understand a word is to know and be able to work with the underlying concepts for which it is an indicator. Understanding between a speaker and a listener occurs when the speaker casts his or her concepts into words and the listener recovers approximately those same concepts. Current models rely on the listener to construct any potential meaning. The diminution of behaviorism as a psychological paradigm and the rise of cognitivism provide examples of many experimental methods that can be used to determine whether and to what extent a machine might understand and to make suggestions about how that understanding might be instantiated.*

*I know there are not words enough in any language to answer all the variety of ideas that enter into men's discourses and reasonings. But this hinders not but that when any one uses any term, he may have in his mind a determined idea, which he makes it the sign of, and to which he should keep it steadily annexed during that present discourse.*

John Locke 1689

Artificial intelligence systems exhibit many useful capabilities, but as has often been said, they lack "understanding," which would be a critical capability for general intelligence. The transformer architecture on which current systems are based takes one string of tokens and produces another string of tokens (one token at a time) based on the aggregated statistics of the associations among tokens. Transformers are purely performative. The representations mediating between the inputs (e.g., prompts) and their production is one purely of the statistical relations among the word tokens. In the case of large language models, we know these facts to be true because this is how the models were designed and they were trained on a kind of fill-in-the-blank test to guess the next word.

What exactly would it mean for an artificial intelligence system to understand? How would we know that it does?

Locke (as quoted above) offers a common argument.  That language tokens are signs for ideas.  Put more generally, this approach argues that the speaker has some idea in mind, selects tokens conditional on that idea.  The listener receives those tokens and selects an idea conditional on those words and on the listener's expectations.  Effective communication occurs when the ideas selected by the listener correspond to the ideas in the speaker's mind.  Applied to a computational machine, language tokens are again signs for underlying representations that are more abstract than the language tokens themselves.  Corresponding to human ideas would be conceptual representations.  Among the properties of these representations are the ability to explain, predict, and entail.  Another way of saying this is that the conceptual representations are what the tokens are about.

Faithful transmission from speaker to listener is often blocked by the fact that there are multiple ways of expressing the same idea (synonymy) and the same tokens can correspond to multiple ideas (polysemy).  A famous example of the ambiguity of language is found in the Delphic Oracle's prophecy to Croesus "if King Croesus crosses the Halys River, a great empire will be destroyed." Croesus understood this advice as a prophecy that he would destroy the Persian army in the Battle of Pteria (547 BC), but it was Croesus's great empire that ended up destroyed (Northwood, 2023).

I got into this work because, sometime in the latter half of the 1980s, I noticed that the word "strike" had 82 definitions in my dictionary.  The number of dictionary definitions a word has is a crude measure of how ambiguous it is, but even that crude measure suggests that the words are only loosely related to their supposed ideas.  Conversely, phrases like "the boy," the "young man," "the youth" and others contain different words, but all mean roughly the same thing.  In a medical record, a patient might be said to "be feverish," "have a fever," "have an elevated temperature," or "have a temperature of 104."

Many other phenomena, including the Tip-of-the-tongue phenomenon (Brown & McNeill, 1966) also indicate a separation between thoughts, representations, and ideas versus words.

As noted earlier, large language models represent only the statistical relationships among tokens.  Thus, they have only one level of representation, which has been found in other domains to be inadequate.  For example, psychological behaviorism attempted unsuccessfully to have a full account, not just of language, but of all behavioral phenomena with a single, behavioral, level of analysis.

Starting in the 1920s and reaching a peak in the 1960s, behaviorists argued that patterns of behavior were caused by reinforcement, which produced a pairing of stimulus (context) with response, such that the reinforced response probability increased in the presence of the stimulus.  Many behaviorists argued that there were no cognitive processes, others argued that even if there are cognitive processes, these processes play no causal role in behavior.

B.F. Skinner, the de facto champion of behaviorism in the 1950s and 1960s, claimed that human language could also be explained by this stimulus-response-reinforcement process. In his view, a person said "Mozart" in response to hearing a symphony because the person has been rewarded in the past for saying that in that context (Skinner, 1957). Noam Chomsky (1967; originally, 1959) called this approach play-acting at science, and, one can argue, was a major impetus for behaviorism's downfall.

On the surface, Chomsky's critique is correct. Reinforcement is not enough to predict what people will say. In hindsight, however, his own position is equally deficient. He argued that human language must require a virtual "language organ," a hypothetical structure in the brain. In his view, language experience set the parameters of this language organ to be consistent with the brain's linguistic milieu, and rejected the notion that language was learned, excepting the specific words of a language.

Chomsky reflected the effort at the time to try to show just what was necessary to explain human behavior. In the case of the transformer models, on the other hand, we know exactly what is sufficient to produce its language patterns and the question is, how can we show whether cognitive processes are necessary to explain it? Is a language model sufficient to produce intelligence?

Unlike with human minds, we know exactly how large language models were constructed. They contain only the relationships between stimuli and responses, aggregated over many instances. Larger language models include larger representations (more parameters) of larger stimuli (contexts) and more associations (parameters) among the elements of the stimulus to make better predictions of the behavior (the words produced). The context and the relations make these models very effective at providing fluent language behavior that mimics human speech and that appears consistent with human cognitive processes, but it is all produced by this basic stimulus-response kind of mechanism. Therefore, it will take extraordinary evidence to convincingly show that large language models contain deeper cognitive processes, or any cognitive processes at all. Despite claims to the contrary (Wei, et al., 2022), evidence of these deeper cognitive processes is so-far missing (Schaeffer, Miranda, & Koyejo, 2023).

The transition from behaviorism to cognitivism following the Chomsky-Skinner debate concerned the very questions that are currently being asked about large language models. Are cognitive processes necessary to explain the observed behavior? In the case of biological entities, the behavioral evidence is available, but the underlying mechanisms is unknown. In the case of GenAI, the underlying mechanism is known and the question is whether the observed behavior requires emergent properties, such as understanding, theory of mind, reasoning, and so forth. Behaviorism tried to reduce apparent cognitive mechanisms to behavior, some GenAI investigators are trying to impute "emergent" cognitive mechanisms. For those challenging behaviorism, the question is whether cognitive processes are necessary. For those challenging artificial intelligence, the question is whether cognitive processes are justified.

# Intervening variables, hypothetical constructs, and latent variables.

There is a lot to learned from the critiques of pure behaviorism as a psychological paradigm. A great deal of analysis over many years went into the question of the necessity of cognition and that analysis is equally applicable to contemporary concerns about large language models and GenAI.

MacCorquodale and Meehl (1948), for example, discuss the difference between *intervening variables* and *hypothetical constructs* in the context of behavioral psychology.  In their analysis, intervening variables are abstractive in the sense that they summarize a set of observations, whereas hypothetical constructs are thought to be entities (constructs) or processes that are so far unobserved, but are thought to exist.  They cite as an example of an intervening variable, the notion of *resistance* in electricity.  This notion summarizes the relationships among volts, amps, and conductors.  They mention *electrons* as an example of a hypothetical construct. Electrons are thought to be particles, and thus objects, but resistance is just an abstractive property. Electrons do not have the property of resistance, but they explain it.  A conductor has the property of resistance, but it is just a summary of, for example, the voltage drop from one end of the conductor to another.  This relationship is expressed by Ohm's law:   V = IR.  I is current, V is voltage and R is resistance.  A law summarizes the relationship among observed variables without explaining the relationship or why that relationship and not another should be observed.  Scientific laws are intervening variables defined by the terms of the observation.

Latent variables are common in contemporary computer science and narrow artificial intelligence, for example, in hidden Markov models and latent Dirichlet allocation and in every application that involves eigenvectors or other forms of matrix factorization.  These latent variables capture the statistical properties of the phenomena in which they are used.  No one credible posits that there is an actual hill inside a computer program that implements gradient descent.  Even the "neurons" of a computational neural network are simply abstractions of a mathematical relationship between their inputs and outputs.  They are metaphors for biological neurons, but they are not thought actually to be entities.

The resistance of a conductor is defined as the ratio of voltage across it to current through it. The voltage drop is not *explained* by *resistance* because resistance is defined by that drop. Analogously, we could define "intelligence" as that which is measured by intelligence tests. Intelligence, under this definition, is not a thing, it is a mathematical relationship that captures the correlations observed among multiple intelligence tests.

The distinction between intervening variables and hypothetical constructs is mostly missing in artificial intelligence thinking.  For example, on the one hand (in the same article), reasoning is defined as: "a cognitive process that involves using evidence, arguments, and logic to arrive at conclusions or make judgments."  And "However, despite the strong performance of LLMs on certain reasoning tasks, it remains unclear whether LLMs are actually reasoning and to what extent they are capable of reasoning."  The former is a hypothetical construct.  The latter use is as an intervening variable, "not actual 'reasoning.'" Is reasoning a thing or is it an expression of the statistical relationship between inputs and outputs? Does reasoning explain the output of a language model (a hypothetical construct), or are the statistical relations extracted from large numbers of texts (an intervening variable) sufficient? (Huang & Chang, 2023)

If reasoning is defined as the performance of a system on reasoning tests (an intervening variable), then reasoning cannot explain the performance on that test any more than resistance explains a voltage drop. It makes no sense to explain a phenomenon by the phenomena that define it. On the other hand, resistance, itself, can be explained in terms of the collisions between electrons and the atoms that make up the conductor. Materials differ in the proportion of electrons that can move freely between atoms, and those that cannot. The lower this ratio, the higher the resistance. We can say that we understand electrical resistance when we can explain it in terms other than the thing we are explaining. Being able to repeat explanatory language similar what has been ingested is not the same thing.

Skinner worked extensively with pigeons (Ferster & Skinner, 1957) to uncover the laws of behavior that governed reinforcement learning. When a pigeon receives food after it pecks a lighted button, a Skinnerian behaviorist might say that an association has been formed between the button and food, but that "association" is nothing more than a description. A cognitivist might claim that the pigeon has learned that the light predicts food or that the pigeon expects food after pecking the lighted key.

In more human terms, consider two people claiming to be "lawyers." One passed the bar exam by memorizing the answers to bar exam questions and one passed by studying and practice. They both exhibit the same behavior, that is, getting the exam answers correct, but only one can claim that they have knowledge of the law.

Large language models are trained to predict the next word given a context of preceding words. This training process models the statistics of word probabilities in the same way that Skinner's pigeons learned to predict the probability of food. As Skinner claimed, they do come to the equivalent of saying "Mozart" following a description of a "Mozart symphony" because they have been reinforced in the past for having produced that response in that context. Reinforcement here means that the model weights were adjusted by the learning algorithm to make the production of Mozart more probable given that context. The model learns a lawful relationship between contexts and productions. These lawful relations are the basis for "distributional semantics" according to which the meaning of words is specified by the context in which they are used (e.g., Harris, 1954). Words in similar contexts tend to have similar meanings.

Language statistics alone are not enough for understanding. The meaning of words is predicted by the words around them, but it is not determined by them. Meaning, in the sense of distributional semantics, is an intervening variable, not a hypothetical construct. A word is selected based on its statistical relation to the words around it, so it cannot be explained by the context used to select it any more than electrical resistance explains the voltage drop across a conductor.

(1) After struggling through three quarters, by the end of the game, they had won.
(2) After struggling through three quarters, by the end of the game, they had lost.

If the two preceding sentences were the complete training set, then distributional semantics would imply that "won" means something very similar to "lost." They both follow the exact same context. Because the large language models are trained on a huge collection of text, the models may have opportunities to learn more relations than are contained in these two sentences. But they would still follow this kind of distributional pattern.

The following is an example (not original) showing that ChatGPT is controlled by this distributional pattern. I provided ChatGPT with the following question:

> Q: "Which weighs more, a kilogram of feathers or a pound of lead?"

**ChatGPT replied:**

> "A kilogram of feathers and a pound of lead both weigh the same, which is approximately 1 kilogram. The difference lies in their units of measurement.
>
> "A kilogram is a metric unit of mass, and a pound is an imperial unit of mass. However, 1 kilogram is equivalent to approximately 2.20462 pounds. So, while they have different units, their weights are equivalent."

The fact that large language models produce mistakes and so-called hallucinations is well known. The lead and feathers example is clearly inconsistent with the idea that the models reason, but it is even more interesting to analyze the example for how the model fails. We can learn more about how a model works by analyzing its failures than its successes.

I did a few Google searches to estimate the frequency with which different word patterns might have been observed in ChatGPT's training set.

4,750 results for "Which weighs more a pound of lead or a pound of feathers?"

1,240 results for "Which weighs more, a pound of feathers or a pound of lead?"

7 results for "Which weighs more, a kilogram of feathers or a kilogram of lead?"

1 result for "Which weighs more, a kilogram of feathers or a pound of lead?" (Posted March 13, 2024 without an answer, long after ChatGPT was trained).

Apparently, the model followed the overall language pattern in this case. The similarity between the language in the training examples and the question I submitted controlled the erroneous output that it produced, not the meaning of the words.

Another observation that is useful for understanding what these models do is the claim that large language models, and not smaller ones, spontaneously learn to add three-digit numbers. Larger language models produce more correct solutions than smaller models. But the mistakes of the smaller language models reveal that as the size of the model increases, the number of digits that they get correct increases (Schafer, et al., 2023, who measure accuracy using token edit distance). For example, for the problem 123 + 456, the correct answer is 579. Token edit distance would assess that 479, 519, and 576 would all be equally partially correct answers. But none of those answers suggest that the model knows anything about multi-digit addition. As large language models grow in their number of parameters, they also grow in the number of tokens used to train them. What appeared to be an scalar emergent property of the number of parameters could actually be a memorization property with more opportunities to include examples of adding 3-digit numbers.

There is an algorithm for adding multidigit numbers (line up the columns, add each column, carry the sum greater than 9) One could use this algorithm without understanding it, but a machine or 3rd grader that operated according to that algorithm could add any multi-digit number, not just up to 3 digits. The fact that it gets some digit tokens wrong is a different pattern than a person using the traditional carry algorithm would make, apparently because the model treats each digit as a token and it fails in the same way that ChatGPT fails the pound of feathers problem.

Finally, a theory of exclusive distributional semantics would view words as defined by the context in which they are used. On this view, the meaning of a word **is** the context in which it appears. *Meaning*, then, would be an intervening variable, capturing the statistical patterns of words given a context. By this definition, each output produced by a large language model would, by definition, be correct, because it conforms to the distributional statistics of its training set. The fact that humans detect errors and hallucinations means that people have information about the meaning of productions that are not contained in the word distributions. A hallucination is at least as statistically valid as a more *correct* production would be in that context, yet we recognize that it is wrong relative to what the observer intended. It cannot be right or wrong, true or false, because its meaning is the probability distribution.

In distributional semantics (taken to the extreme) the word is just a statistical regularity appropriate to the context, like resistance in a wire. Identifying the additional semantic content, its source, and its representation is the topic of the rest of this essay.

## Positivist semantics and theories

Scientific thinking is done (so far) by human scientists. These people are not much different from other humans and their thinking is not much different from other kinds of thinking (except maybe for the topic). To the extent that there is a difference, it may be in that some scientists, supported by philosophers of science, do their thinking more deliberately and pay more attention to the processes that they employ (Phillips, 1990; Creath, 2023), than do other people.

The logical positivists were a group of philosophers centered around Vienna, Austria in the early 20th Century, (1920-1930s). The New Physics of relativity and quantum mechanics was emerging at this time, radically changing scientists' views of Newton's mechanics, and of the nature of scientific thought. Euclidean geometry, laws of motion, simultaneity, and many other terms required radical reinterpretations. The logical positivists sought to define the nature of meaning to avoid the need for such future reinterpretations. They declared that for a sentence to be *meaningful*, it must either be verifiable by observation(s) or be a logical consequence of verifiable observation(s).

Propositional logic statements transmit truth from their premises to their conclusions. If the premises are correct and the statement is of the correct form, then the conclusion will also be correct. Extreme distributional semantics is a similar attempt to reduce meaning to formal properties of statements, just with less formality. For positivists, the meaning of a term is the propositional logic in which it is used, in extreme distributional semantics, it is the probability distribution of the terms with which it is used.

The positivist claim, then, provides a strong potential guide for machine understanding. Logic alone is not enough to be meaningful, but logic plus observation might be. Propositional logic transmits truth from the premises to the conclusion. The logical statement: "All x are y. A is an x, therefore, A is a y," is of a proper logical form, but it cannot be said to be correct or incorrect unless we know what A, x and y are. The positivists relied on observation to tell them what A, x, and y were. In large language models there is no similar method.

If we replace A with "Daneel," x with "man," and y with "mortal," then we have a statement that could be true or false, depending on our observations. It is, therefore, meaningful. Meaningful statements are those that could be right or could be wrong.

Even for the logical positivists, meaning required observation that is different from the symbols contained in the statement. Is Daneel a man? Are men mortal? If Daneel were, say, an immortal robot, then the sentence would be false because our observations were inconsistent with the first premise, but the logical formula would be valid. The observations could be said to ground the statement (symbol grounding will be considered shortly).

The positivist program failed in part because the meanings of scientific terms (all terms actually) cannot be reduced to observations and deductions. (Phillips, 1990)

Among the reasons that it failed are:

- Observations are not infallible. Observations can be mistaken.
- Deductions carry truth, they do not increase it.
- Terms can be meaningful even if we cannot specify the criteria for making them true.
- Scientific statements may refer to observations that have not yet been made.
- Scientific statements transcend the observations that have been made.

Natural language is ambiguous, as discussed earlier, but even if that ambiguity were eliminated, for example, through use of a strictly defined symbolic logic, it still would not capture meaning. Unambiguous sentences can carry truth from certain kinds of sentences to others (from premises to conclusions), but they carry that truth exactly as well if the premises are meaningful as they do when the premises are not meaningful.

Language models learn statistical patterns that connect premises to conclusions. The statistical patterns ensure that the model's productions are sound in the sense that they are samples from the token patterns observed during training, but without the guarantees of propositional logic, they cannot even guarantee that the conclusion, their output, follows from the input, merely that it is statistically probable. Because the premises, the context, has no meaningful relationship to anything but other symbols, they cannot guarantee that the output is even meaningful, let alone true.

The failure of the positivist program led philosophers of science to take the tack of eliminating the positivist quest for certainty. The positivists thought that they could make statements that were meaningful and certain.  Post-positivists, sacrifice certainty in order to be meaningful.  Meaningful sentences are grounded in observation, but they are not guaranteed by the observation.  Scientists can judge some statements as being more meaningful than others in the context of observations, but these judgments are subject to revision with further observation.

## The symbol grounding problem

Similar distributional patterns can identify that two words, for example, have similar meanings, but not what those meanings are.  The positivists grounded the symbols in scientific statements by connecting them to observation statements.  A later view of the symbol grounding problem was described by Harnad (1990) in the context of Searle's (1980) Chinese Room thought experiment.  Searle asked his readers to imagine a room filled with baskets of tokens, each of which represented a Chinese character.  The room contained a window through which an outsider could pass tokens and another window through which the occupant of the room could respond with other tokens.  Additionally, the room contained a rule book that described in perfect detail the appropriate tokens to push out the window for every group of tokens pushed in.  The rule book is the equivalent of the language model, but it would also have to have properties that the language does not.

The person in the room could read the book, but knew no Chinese, only the rules for accepting and producing tokens.  To the person outside the room, the tokens going into and coming out the room represented a perfect Chinese conversation, but no one in the room knew Chinese.  Searle asserted that this thought experiment meant that a computer, which knew only rules, could appear to hold a conversation without knowing at all what the conversation was about.  The computer knew only the syntax (the rules in the book) and had no access to the semantic properties of the language.

In Harnad's words, the meaning of the conversation as perceived by the Chinese speaker outside the room was "parasitic" on the speaker's knowledge.  The person interpreted the room's symbols as meaningful, because of the person's knowledge of Chinese.  The symbols, by hypothesis, did not have any meaning, just the rule book.

This is the symbol grounding problem: "How can the semantic interpretation of a formal symbol system be made intrinsic to the system, rather than just parasitic on the meanings in our heads?" The problem, he said, was like how to learn Chinese from a Chinese only dictionary.

How can we get from systems of meaningless symbols, like a large language model, to meaning? Searle's answer was that we can't but his reasoning was arbitrary.  Only brains, specifically brains of a certain type held only by humans so far as we know were capable of having semantics. Computer programs could not have this ability.

In Harnad's view, meaning required two kinds of nonsymbolic representations, either analogs of the sensory properties of objects and events, or categorical representations of the invariant features of objects and events.  Elementary (base-level) symbols are the names of the object and event categories, and higher-order symbols consist of combinations of these base-level symbols.

The critical idea for large language models is that meaning requires connection to information that is not contained in the language tokens and their relationships (Locke's "ideas").  From the perspective of the language model, every production that conforms to the learned statistical model is equally valid.  For the symbols to have meaning, they must also be constrained by outside factors that nonsymbolically represent categories of invariant patterns of the world.

Nvidia CEO Jensen Huang was recently quoted as saying that the hallucinations, common for large language models, are easy to cure by adding a rule: "For every single answer, you have to look up the answer."  https://techcrunch.com/2024/03/19/agi-and-hallucinations/ Prevent the model from producing any text that cannot be verified by other text.  In that same presentation, he also claimed that artificial general intelligence would emerge within the next 5 years, but he had a peculiar definition of general intelligence as "a set of tests where a software program can do very well — or maybe 8% better than most people."  This prediction is saved from complete internal contradiction by redefining general intelligence narrowly as the passing of specific tests. He simply defined the problem of general intelligence out of existence.  No understanding is required for a computer to pass these predefined tests, but there is still the contradiction that his purported intelligence would be limited to just those things that that can be verified as having been said before.  Generative AI is reduced to a search engine, and artificial general intelligence is reduced to (defined as) passing some tests.

An essential feature of intelligence is to be able to say things that have not been said before, that reveal new facts, insights, or relations that have so far not been apparent.  Intelligence invents.  It *understands* in new ways.  The best-described examples of invention involve science or math, but invention occurs in more quotidian circumstances as well.  Resistance, as described earlier, is explained by the proportion of free electrons.  Knowing about electrons (a hypothetical construct) can help to explain why light shining on certain materials can create an electric current.  It turns out that the effect of light requires an additional understanding to view light as particles (photons) interacting with the electrons.  But before Albert Einstein wrote about this understanding, it could not be found anywhere (Elert, 2024).  It could not be verified, therefore, by other texts.  In fact, it was contradicted by much of the writing that preceded him, but it was not a hallucination, and eventually this insight won him the Nobel Prize.  The words that he used were undoubtedly widely known.  Even many of the phrases were probably known.  His conclusion was not.

The mathematician, physicist, and philosopher, Henri Poincare, described his thought processes during mathematical invention.

> "In fact, what is mathematical creation? It does not consist in making new combinations with mathematical entities already known. Any one could do that, but the combinations so made would be infinite in number and most of them absolutely without interest."

The interesting combinations "are those which reveal to us unsuspected kinship between other facts, long known, but wrongly believed to be strangers to one another."

> "For fifteen days I strove to prove that there could not be any functions like those I have since called Fuchsian functions. I was then very ignorant; every

*day I seated myself at my work table, stayed an hour or two, tried a great number of combinations and reached no results. One evening, contrary to my custom, I drank black coffee and could not sleep. Ideas rose in crowds; I felt them collide until pairs interlocked, so to speak, making a stable combination. By the next morning I had established the existence of a class of Fuchsian functions, those which come from the hypergeometric series; I had only to write out the results, which took but a few hours."* (Poincare, 2013; Wagenmakers, 2022)

I am not concerned here with the psychological aspects of Poincare's invention process, the black coffee, the lack of sleep, or the suddenness of the insight. What I do find interesting in the present context is that the insight depended on certain combinations of concepts. The combinatorics of trying every pair would make a random search impractical, even for a very large computing footprint. Either Poincare was very lucky to hit on interesting combinations during his lifetime (he reports doing it repeatedly) or he had some other, still unknown, method of selecting interesting candidates. For example, perhaps some kind of diffusion method would serve to nonrandomly select candidates. It would be very useful to have such a method.

Another important point for our purposes is that he could not or did not succeed in describing this relationship prior to his discovery. Once he identified the inventive relation, he could describe it, but those descriptions were not available before then. The process of invention was, apparently, separated from the process of description, it appears not to have been mediated by language patterns.

Harnad argued that symbols were grounded by three kinds of concepts: sensory representations, categorical representations, and combinatorial representations (my names). The first two are derived from raw sensory experience or from detection of invariant properties of sensory experience. Sensory representations would be the equivalent of pixel-level representations of visual scenes. Categorical representations would be the equivalent of object detection (perhaps among others). But there is more to meaning, and thus to understanding, than could be provided by sensory experience.

Many concepts, such as *goodness*, *truth*, and *beauty*, do not correspond directly to any specific sensory experience. An apple detector may detect invariant features of apples, but that does not imply that the symbol *apple* **means** the objects the detector picks out or the process of picking them out. If the meaning of *apple* was the set of all apples in the world, then the meaning would change with each new crop. Conversely, one could believe that apples are good to eat and not believe that a wax apple (with an appearance very similar to a delicious apple) was good to eat. You might think that a crab apple was good to eat, but not a road apple.

The members of a category may have no sensory features in common. Wittgenstein discussed categories such as games (e.g., solitaire, chess, football, tag) and furniture (e.g., sofas, tables, lamps) without shared sensory features. These would have to be conceptual categories.

Ultimately, the conclusion is that most or maybe all concepts do not have invariant sensory features. Nor is their meaning contained in reference to specific objects or activities. Rather, a more useful approach is probably to consider that the core concepts resemble folk theories, similar to scientific theories. Folk theories are usually implicit and less formal than scientific theories. They concern

> *how the mind interprets new information and constructs new ways of understanding. We propose that mental content can be productively approached by examining the intuitive causal explanatory 'theories' that people construct to explain, interpret, and intervene on the world around them*
> (Gelman & Legare, 2011)

Folk theories model the concept they represent rather than attempt to be some faithful copy of sensory experience. The logical positivists tried to remove such entities from science so that every scientific statement could be verified. They failed in part because the verification of theoretical constructs depends on prior theoretical commitments and is ultimately impossible. There is no reason to think that verificationist theories of concepts should be immune from the problems that killed positivism. Rather, science, and the theoretical view of concepts, accept the fact that these representations are uncertain and subject to revision. As the statistician George Box noted, all models are wrong, some are also useful, to which we might add, none are guaranteed.

On the view described here, understanding means connecting sentences or other text units to theory-like concepts. Like scientific theories, these concepts are derived from linguistic and non-linguistic experience. How exactly they are constructed has yet to be investigated, but the key idea is that they are separate from the statistics of language and imperfectly related to the world and to the units of language. They are not just intervening variables capturing statistical relations, but are thought to be actual entities, that is, hypothetical constructs.

Again, like scientific theories, concepts transcend, abstract, and extrapolate the specific objects and events from which they are derived. Take a small child to the zoo and buy her a cotton candy and she will create a model that predicts cotton candy on all future zoo visits.

This view of understanding takes no position on sentience, consciousness, or even on cognition. It also takes no position on whether "embodiment," that is, having the experience of a body, or having a biological brain is relevant. It does take the position that language statistics are not enough to produce understanding and that the semantics of language is the relation between the units of language and these proposed theory-like concepts. Understanding, then, is something that computers may accomplish, but research will be needed to accomplish it.

The properties of these theory-like representations should, I think, be the target of immediate research. We can gather some characteristics that are likely to be important from studies of human folk theories and categories. Intelligent machines do not have to solve problems in the same way that people do, but they would have to solve most of the same problems.

These properties might include:
- contextually modulated similarity judgments (similarity changes with context)
- representations of objects (objects, not just pixel patterns; object permanence)

- representations of causal relations
- entailment relations (Schroeder is 9th Chancellor implies that 9th Chancellor is Schroeder)
- physics of motion
- representation of meaning (not just token patterns)
- typicality structure for categories

Many of these have been studied in humans, so there are many clues for how to start investigating them in machines. Some sources include (Rips, Shoben, & Smith, 1973; Rosch, 1973; Smith & Medin, 1981; Barsalou, 1983; Gentner, 2003)

Investigating machine understanding will require careful experimental methods. Many of the current investigations that seek to celebrate the "emergence" of some cognitive process depend on observing a success at solving a problem that supposedly depends on the presence of that cognitive process. This method corresponds to the logical fallacy of "affirming the consequent." I think that Mickey stole the cookies from the cookie jar and my evidence of this is that the cookies are missing. This form of argument neglects that someone else may have taken the cookies. There could be other explanations.

When the models get a problem right, there could be several explanations, including that they have been exposed to that linguistic pattern during training. When they get problems wrong, on the other hand, we can often learn a lot from how they got it wrong. To paraphrase Tolstoy, happy hypotheses are all alike, but each error is unhappy in its own informative way. A general heuristic for error analysis is that similar representations are more likely to be confused with one another than dissimilar representations. The point is not that models make errors, they do, but the patterns of those errors can be revealing.

By this light, computational models are, in principle, capable of understanding. This essay lays out some of the minimal criteria by which we may build and evaluate such models. It would be a mistake, however, to take these criteria as operational definitions of understanding, as tests to be passed. Other models may be able to meet these criteria without understanding.

It would also be premature to conclude that these are definitive criteria. Rather, they should be taken as hints as to what to look for and how to look for it. I intend this essay to mark the beginning of a research program that will help to move past stochastic parrots to models that show at least rudimentary understanding.

A key idea of this essay is that understanding is potentially within the capabilities of machines. It does not require biological brains, sentience, consciousness, enlightenment, or any other ineffable properties. Achieving understanding will not lead inevitably to super-intelligence but it will probably lead to more efficient, effective, and more general artificial intelligence.